\documentclass{article}

\usepackage{PRIMEarxiv}

\usepackage{amsmath}
\usepackage[utf8]{inputenc} 
\usepackage[T1]{fontenc}    
\usepackage{hyperref}       
\usepackage{url}            
\usepackage{booktabs}       
\usepackage{amsfonts}       
\usepackage{nicefrac}       
\usepackage{microtype}      
\usepackage{lipsum}
\usepackage{fancyhdr}       
\usepackage{graphicx}       
\graphicspath{{media/}}     

\title{EvoMerge: Neuroevolution for Large Language Models
}

\author{
  Yushu Jiang \\
  \texttt{barryy.jiang@mail.utoronto.ca} \\
}

\begin{document}
\maketitle

\begin{abstract}
Extensive fine-tuning on Large Language Models does not always yield better results. Oftentimes, models tend to get better at imitating one form of data without gaining greater reasoning ability and may even end up losing some intelligence. Here I introduce EvoMerge, a systematic approach to large language model training and merging. Leveraging model merging for weight crossover and fine-tuning for weight mutation, EvoMerge establishes an evolutionary process aimed at pushing models beyond the limits of conventional fine-tuning.
\end{abstract}

\section{Introduction}
In this paper, I propose the idea that we apply neuroevolution methods as a system for training large language models. Specifically, we use model merging algorithms for weight crossover and reproduction, and fine-tuning as weight mutation. So far this is less of a technical paper where I show my findings, but rather, to get the idea out there and encourage further experimentation in the community.\\\\
\textbf{Important: This is the first draft, published for the sole purpose of sharing an idea and encouraging community effort. A more consolidated version may come later. (Please forgive the crappy writing.)}

\section{The Idea}
The inspiration behind this paper mainly comes from popular neuroevolution algorithms such as NEAT \cite{stanley:ec02}, as well as the surprisingly well-performing series of Mistral merges such as mlabonne/NeuralBeagle14-7B \cite{mlabonneNeuralBeagle14}. NeuralBeagle14-7B has shown us that simply merging and tuning with DPO repetitively improves the model further and further. So what if we keep doing it? We train a series of variants, merge them together, and we do another round and merge them again. Will the model keep getting better or will it hit a limit? My intuition tells me that the madness is going to stop somewhere. However, when that happens, what if we then use a different set of hyper-parameters, switch to a different dataset, or use a different training method? Surely there is more to learn and we definitely haven't tried everything yet. Now, I don't have the answer to any of those questions. But what I do have is a potential way to answer those questions someday, and in the meantime, improve your models further. This is, the "try everything" method. Instead of training one model at a time, we train multiple of them, with different parameters, datasets, methods, etc, we can try everything. Then, we evaluate them, to see which ones are doing better. And finally, we follow the rule of the universe, "survival of the fittest". Since better weights are more likely to survive and reproduce, I would expect that only beneficial training runs with be kept and the model will continue to develop better activations.
Now that, was the big idea. Evolution for large language models. In the following paragraphs, I will provide a general design of such an evolutionary system. Keep in mind that there should be better designs, and this is simply my intuition.

\section{The Steps for Evolution}
Taking inspiration from neuroevolution algorithms, there should be 6 key steps: Initialization, Evaluation, Crossover, Selection, Mutation, and Repeat.
\begin{figure}[hbt!]
    \centering
    \includegraphics[width=1\linewidth]{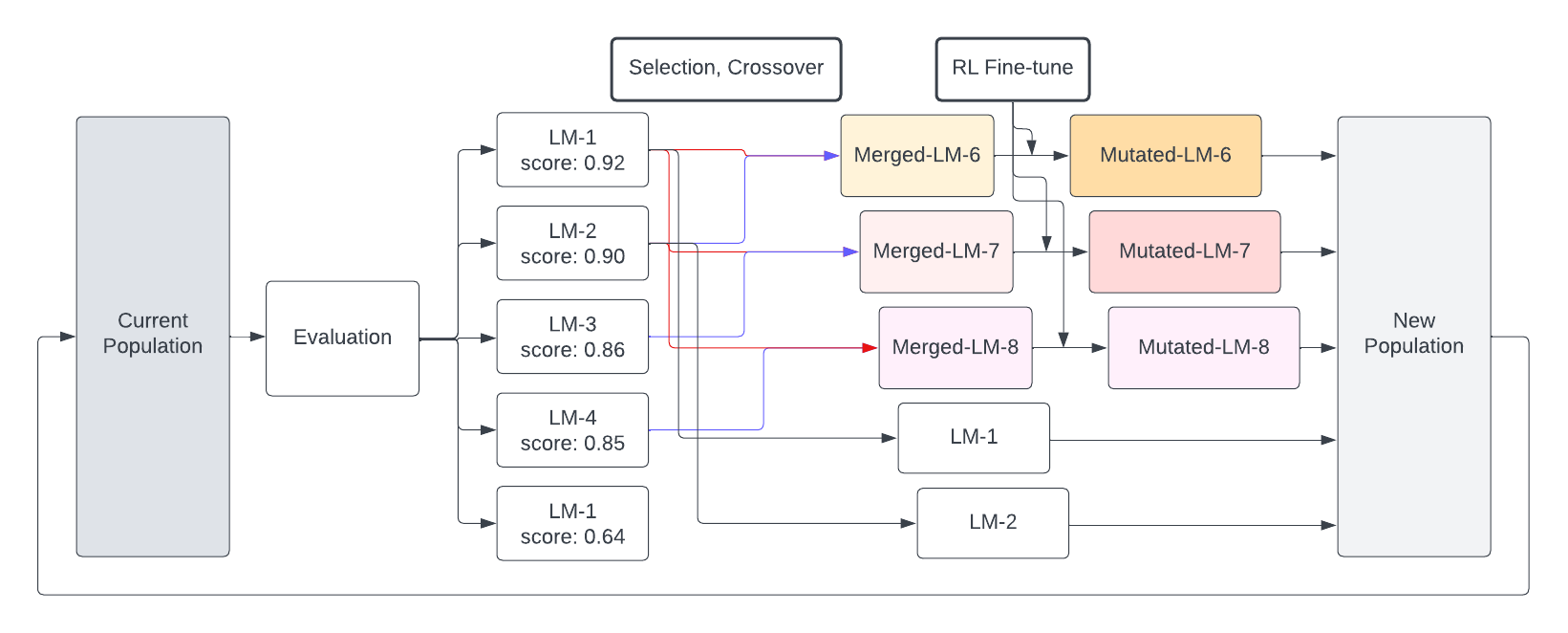}
    \caption{Evolution loop}
    \label{fig:}
\end{figure}

\subsection{Initialization}
The initialization stage involves creating the first batch of candidates. There are not many rules here, further research needs to be conducted. Intuitively, however, a higher-quality initial population should yield better results or at least speed things up in the beginning.

\subsection{Evaluation}
The evaluation stage involves using a set of evaluation methods to determine the fitness of each model, in other words, this stage determines how good each model is. There can be many potential evaluation methods. However, we are always at risk of having models overfitting on the fitness function. For example, evaluating with math problems does not tell us whether or not the model is improving in any other aspects even if the datasets contains general information. Evaluation is crucial to ensure incremental improvements, thus needs to be designed carefully. Further research can be done here.

\subsection{Selection}
The selection stage is responsible for planning the next generation of models based on the fitness scores of the current generation.  The new generation will consist of recombined individuals and the old generation will be deleted. In theory, we want better-performing models to reproduce and create new generations of better models. However, to prevent loss of diversity and take into account the inaccuracy of evaluation methods, some randomness should be added when selecting pairs for reproduction. Furthermore, in the hope of reducing instability in the evolution process where all candidates drop in performance leading to worse generations down the line, we can keep the top k best-performing models and prevent them from mutating. Other stabilization methods might include keeping multiple generations, further research needs to be conducted.

\subsection{Crossover}
The crossover stage takes the picked pairs/groups from the selection stage, combines them, and creates the next generation. For evolving model weights, we can use popular model merging methods such as Spherical Linear Interpolation (SLERP), TIES \cite{yadav2023tiesmerging}, or DARE \cite{yu2023language}. Since there are quite a handful of parameters we can tinker with, evolving parameters as genes can be useful. However, such a method requires additional design and research beyond the scope of this paper. Additionally, more research can be done to allow crossovers of LLMs with different bases.

\subsection{Mutation}
The mutation stage involves fine-tuning reproduced models to introduce variations to model weights. The tuning method can vary, though, the WARM \cite{ramé2024warm} paper has shown that reward models benefit from weight averaging. Through mutations is how we make sure that we are not just getting a weight average of the initial population. With each mutation, the model has a chance to become better than its parents. Furthermore, intuitively, bad training runs with either bad data or hyperparameters will be filtered out through evolution. With that said, it is also possible that low-quality training could destabilize the evolution process. Further study is required to find a better range of randomness. One other thing that should be interesting to look into is the benefit of high-quality data and how much further we can push with just one single high-quality dataset.

\section{EvoMerge: The Prototype}
The EvoMerge prototype is a small-scale experiment to showcase some potential benefits of adopting a neuroevolution system for improving large language models. The project is ongoing and I will be updating this paper if any valuable results come up.\\
The overall design:
\begin{itemize}
\item Randomly sampled from HellaSwag \cite{zellers2019hellaswag} and WinoGrande \cite{sakaguchi2019winogrande} for \textbf{evaluation}
\item Roulette wheel \textbf{selection}
\item Spherical Linear Interpolation (SLERP) for \textbf{crossover}
\item DPO \cite{rafailov2023direct} fine-tuning on the argilla/distilabel-intel-orca-dpo-pairs dataset as \textbf{mutation}
\end{itemize}

\subsection{Experiment-1}
Initial population:
\begin{itemize}
\item mlabonne/NeuralBeagle14-7B
\item udkai/Turdus
\end{itemize}

Ran for 5 generations:
\begin{table}[htb]
  \centering
  \caption{Experiment-1 Result}
  \label{tab:llm-scores}
  \begin{tabular}{lccc}
    \hline
    \textbf{Language Model} & \text{NeuralBeagle14-7B} & \text{Turdus} & \textbf{EvoMerge Result}  \\
    \hline
    Average & 74.74 & 72.25 & \textbf{74.83} \\
    ARC & \textbf{73.38} & 72.95 & 73.12 \\
    HellaSwag & 88.56 & 88.34 & \textbf{88.61} \\
    MMLU & 64.52 & 64.55 & \textbf{64.75} \\
    TruthfulQA & 67.11 & 69.93 & \textbf{69.99} \\
    Winogrande & ? & 82.4 & \textbf{85.16} \\
    GSM8K & 67.7 & \textbf{70.28} & 67.32 \\
    \hline
  \end{tabular}
\end{table}

\subsection{Experiment-2}
Initial population:
\begin{itemize}
\item BarryFutureman/NeuralTurdusVariant1-7B (Result from Experiment-1)
\item mlabonne/NeuralDaredevil-7B
\item senseable/WestLake-7B-v2
\item PetroGPT/Severus-7B-DPO
\end{itemize}

Ran for 5 generations, aimed to improve GSM8K score:
\begin{table}[htb]
  \centering
  \caption{Experiment-2 Result}
  \label{tab:llm-scores}
  \begin{tabular}{lccccc}
    \hline
    \textbf{Language Model} & \text{NeuralDaredevil} & \text{WestLake-v2} & \text{Severus} & \text{Experiment-1} & \textbf{EvoMerge Result}  \\
    \hline
    Average & 74.12 & 74.68 & 72.81 & \textbf{74.83} & \textbf{74.81}\\
    ARC & 69.88 & 73.04 & 70.22 & 73.12  & \textbf{73.21} \\
    HellaSwag & 87.62 & 88.34 & 87.09 & \textbf{88.61} & 88.37 \\
    MMLU & 65.12 & 64.55 & \textbf{64.93} & 64.75 & 64.76 \\
    TruthfulQA & 66.85 & 69.93 & 64.41 & \textbf{69.99} & 68.09 \\
    Winogrande & 82.08 & 82.4 & 80.66 & \textbf{85.16} & 84.07 \\
    GSM8K & \textbf{73.16} & 70.28 & 69.52 & 67.32 & 70.05 \\
    \hline
  \end{tabular}
\end{table}

\newpage
\subsection{Experiment-3}
Initial population:
\begin{itemize}
\item BarryFutureman/NeuralTurdusVariant1-7B (Result from Experiment-1)
\item BarryFutureman/WildWest-Variant3-7B (Result from Experiment-2)
\item Toten5/Marcoroni-neural-chat-7B-v2
\item alnrg2arg/blockchainlabs-7B-merged-test2-4
\end{itemize}

Ran for 6 generations:
\begin{table}[htb]
  \centering
  \caption{Experiment-3 Result}
  \label{tab:llm-scores}
  \begin{tabular}{lccccc}
    \hline
    \textbf{Language Model}  & \text{Marcoroni-neural-chat} & \text{blockchainlabs} & \text{Experiment-1} & \text{Experiment-2} & \textbf{EvoMerge Result}  \\
    \hline
    Average & 72.5 & 75.28 & 74.83 & 74.81 & \textbf{75.29}\\
    ARC & 68.6 & 73.55 & 73.12  & 73.21 & \textbf{73.98} \\
    HellaSwag & 86.33 & \textbf{88.87} & 88.61 & 88.37 & 88.61 \\
    MMLU & 64.65 & 64.63 & 64.75 & 64.76 & \textbf{64.81} \\
    TruthfulQA & 61.84 & 69.77 & \textbf{69.99} & 68.09 & 69.76 \\
    Winogrande & 80.43 & 84.45 & \textbf{85.16} & 84.07 & 84.29 \\
    GSM8K & \textbf{73.16} & 70.43 & 67.32 & 70.05 &70.28 \\
    \hline
  \end{tabular}
\end{table}

\section{Conclusion}
Nothing to conclude here, this research is not yet over. Stay tuned for future revisions.

\bibliographystyle{unsrt}
\bibliography{references}

\begin{thebibliography}{1}

\bibitem{stanley:ec02}
Kenneth~O. Stanley and Risto Miikkulainen.
\newblock Evolving neural networks through augmenting topologies.
\newblock {\em Evolutionary Computation}, 10(2):99--127, 2002.

\bibitem{mlabonneNeuralBeagle14}
Maxime Labonne.
\newblock Neuralbeagle14-7b.
\newblock \url{https://huggingface.co/mlabonne/NeuralBeagle14-7B}, 2024.

\bibitem{yadav2023tiesmerging}
Prateek Yadav, Derek Tam, Leshem Choshen, Colin Raffel, and Mohit Bansal.
\newblock Ties-merging: Resolving interference when merging models, 2023.

\bibitem{yu2023language}
Le~Yu, Bowen Yu, Haiyang Yu, Fei Huang, and Yongbin Li.
\newblock Language models are super mario: Absorbing abilities from homologous models as a free lunch, 2023.

\bibitem{ramé2024warm}
Alexandre Ramé, Nino Vieillard, Léonard Hussenot, Robert Dadashi, Geoffrey Cideron, Olivier Bachem, and Johan Ferret.
\newblock Warm: On the benefits of weight averaged reward models, 2024.

\bibitem{zellers2019hellaswag}
Rowan Zellers, Ari Holtzman, Yonatan Bisk, Ali Farhadi, and Yejin Choi.
\newblock Hellaswag: Can a machine really finish your sentence?, 2019.

\bibitem{sakaguchi2019winogrande}
Keisuke Sakaguchi, Ronan~Le Bras, Chandra Bhagavatula, and Yejin Choi.
\newblock Winogrande: An adversarial winograd schema challenge at scale, 2019.

\bibitem{rafailov2023direct}
Rafael Rafailov, Archit Sharma, Eric Mitchell, Stefano Ermon, Christopher~D. Manning, and Chelsea Finn.
\newblock Direct preference optimization: Your language model is secretly a reward model, 2023.

\end{thebibliography}

\end{document}